%% file: main.tex
\documentclass[runningheads]{llncs}
\usepackage{amsmath}
\usepackage{graphicx}
\usepackage{comment}
\usepackage[colorinlistoftodos]{todonotes}
\begin{document}
\title{Self-Driving Car Steering Angle Prediction: Let Transformer Be a Car Again.}
\titlerunning{Team 14}
\author{Chingis Oinar\inst{1}\orcidID{2018315169} \and\\
Eunmin Kim\inst{3}\orcidID{2018315083}}
%

\institute{Sungkyunkwan University\\
2066, Seobu-ro, Jangan-gu, Suwon-si, Gyeonggi-do, Republic of Korea \and
\email{chingisoinar@gmail.com} \and
\email{eunmin88@gmail.com}}

\maketitle              % typeset the header of the contribution
\input{0_abstract}

\input{1_intro}
\input{2_objective}
\input{3_related}
\input{challenge}
\input{4_solution}
\input{4_settings}
\input{5_Planning}
\input{conclusion}
\input{6_references}
%
%
%

\end{document}

%% file: 0_abstract.tex
\begin{abstract}
Self-driving vehicles are expected to be a massive economic influence over the coming decades. Udacity\footnote{https://www.udacity.com/} has been working on a completely open-source self driving car. Thus, it regularly organizes various competitions, one of which was dedicated to steering angle prediction task. In this work, we perform an extensive study on this particular task by exploring the Udacity Self-driving Car Challenge 2. We provide insights on the previous teams' solutions. Moreover, we propose our new architecture that is inspired by some of the teams. We report our performance and compare it with multiple baseline architectures as well as other teams' solutions. We make our work available on GitHub and hope it is useful for the Udacity community and brings insights for future works\footnote{https://github.com/chingisooinar/AI\_self-driving-car}. 

\end{abstract}

%% file: 1_intro.tex
\section{Introduction}
Self-driving vehicles are expected to be a massive economic influence over the coming decades. Convolutional Neural Networks (CNNs) have been widely used in pattern recognition tasks. Moreover, there have been numerous attempts to adopt CNNs beyond the classical pattern recognition by applying it in steering angle prediction tasks. One of the famous attempts was done by NVIDIA that introduced its novel CNN-based architecture, DAVE-2, in 2016 \cite{DAVE2}.  They collected training data by driving on a wide variety of roads and in a diverse set of lighting and weather conditions. Thus, the data was collected from surface street data in central New Jersey and highway data from Illinois, Michigan, Pennsylvania, and New York. Additionally, data was gathered under different weather conditions, including clear, cloudy, foggy, snowy, and rainy weather. Udacity, an online educational platform, has an ongoing challenge to produce an open source self-driving car. In their second challenge Udacity released the exact same dataset collected by NVIDIA in 2016. The goal of the challenge was to come up with an architecture that will minimize the RMSE (root mean square error) between the prediction and the actual steering angle produced by a human driver. In this project, we explore a variety of approaches including LSTM and transfer learning, and we will also propose our new architecture based on Transformer.

%% file: 2_objective.tex
\section{Motivation \& Objective}
Self-driving vehicles have always been a fantasy for people all over the world. With the recent introduction of AI to self-driving vehicles, that fantasy has become a reality. Recently Udacity hosted a challenge where teams can compete for the lowest RSME score. The first place team, komanda, achieved a public score of .0483 and a private score of .0512. Our main motivation rooted from this challenge because we wanted to see if our newly created model would be superior to others in the challenge. Our main objective is to improve RSME over the first place team with a newly created model using optical flow and transformer.

%% file: 3_related.tex
\section{Related Work}
\subsection{Deep Neural Networks for Self-driving vehicles.}
In 2016 NVIDIA came up with an interesting solution by utilizing Convolutional Neural Networks (CNN) to predict steering angles. They proposed a simple CNN-based architecture, called DAVE2 \cite{DAVE2}, which was trained on video frames collected from three independent cameras, simulating different perspectives, in order to imitate a legit human driver. Surprisingly, it achieved good results and was able to maneuver around a city. Recently, there have been more attempts on using deep CNNs and Recurrent Neural Networks (RNNs) to tackle the problems of video classification \cite{video_class}, scene
parsing \cite{scene_parsing}, and object detection \cite{detection}. Thus, this trend has also inspired the applications of more complicated and sophisticated CNN architectures in autonomous driving.

Moreover, there have been various and deeper research initiatives adopting deep learning techniques in autonomous driving challenges. 
For example, Comma.ai has proposed to learn a driving simulator by combining a Variational Auto-encoder
(VAE) \cite{VAE} and a Generative Adversarial Network (GAN) \cite{GAN}. Currently, NVIDIA provides NVIDIA DRIVE Sim\footnote{https://developer.nvidia.com/drive/drive-sim}, powered by omniverse and GAN, which is a scalable, physically accurate, and diverse simulation platform.

Deep reinforcement learning (RL) has also
been applied to autonomous driving \cite{RL_car}. After a tremendous success of Google DeepMind that demonstrated the capability of RL by learning of games like Atari and Go, there have been numeruos attempts to adopt it for autonomous vehicles and different simulators \cite{Atari}. One of them is a The Duckie Town Simulator\footnote{https://www.duckietown.org/} that started as a class at Massachusetts Institute of Technology (MIT) and a lane-following simulator yet later evolved into a fully-functioning autonomous driving simulator.

%% file: challenge.tex
\section{Udacity Self-driving Car Challenge 2}
\subsection{Dataset}
The dataset\footnote{https://github.com/udacity/self-driving-car/tree/master/datasets/CH2} that we chose to use is the Udacity dataset. Inside the dataset, there are three distinct types of images. The first one is named "center" and it is where the car POV is in the center of the car itself. The second one is named "right" and the POV is from the passenger's seat. The last one is named "left" and the POV is from the driver's seat. The images have all different types of road conditions, weather conditions, number of cars on the road in front of them, daytime/nighttime, angle, turn, direction of the sun, and areas. The images were captured in the following areas: Northbound and Southbound on El Camino, and a Lap around the block at the Udacity office. The images were shot using IMU positioning and HDL-32E LIDAR.
Training data set contains 101397 frames, whereas the test data set consists of  5615 frames.
The following three figures shows three example images from the dataset. They all have different settings and environments.
In Figure 1, we can see the image is on a two lane road on a right turn with a left POV. The sunlight is strong so it must be in the middle of the day or morning.
In Figure 2, there is a center POV where it seems like the highway or a busy road with lots of cars within the vicinity. The time of day seems to be around sunset.
In the final Figure 3, we can see a right POV on a road where seems to be taking a left turn with rails on the right side. time of day seems to be midday with no other cars in the vicinity.
\begin{figure}
\begin{minipage}[c]{0.3\linewidth}
\includegraphics[width=\linewidth]{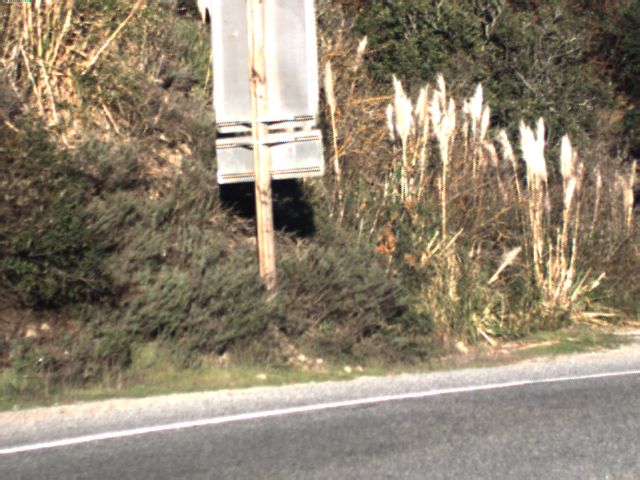}
\caption{Left POV}
\end{minipage}
\hfill
\begin{minipage}[c]{0.3\linewidth}
\includegraphics[width=\linewidth]{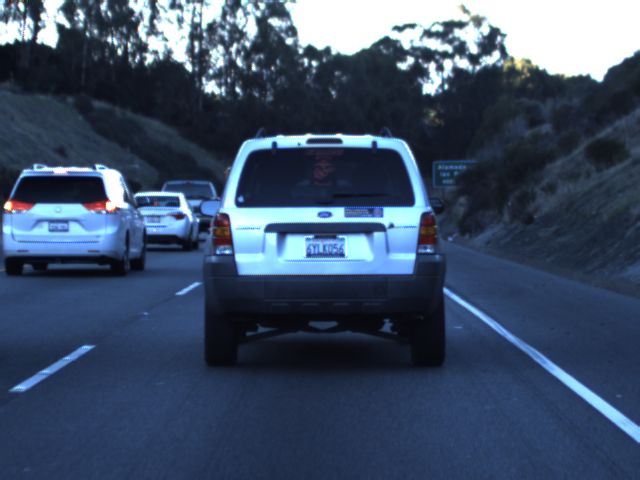}
\caption{Center POV}
\end{minipage}%
\hfill
\begin{minipage}[c]{0.3\linewidth}
\includegraphics[width=\linewidth]{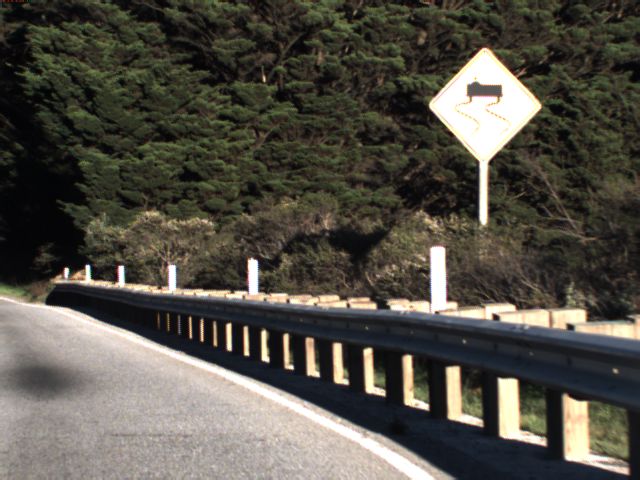}
\caption{Right POV}
\end{minipage}%
\end{figure}
\subsection{Revisiting Previous Solutions}
\subsubsection{Rambo.} There are few things we need to emphasize regarding the team's solution. During the preprocessing stage, the team converted RGB images into grayscale after which they computed lag 1 differences between frames and utilized 2 consecutive differenced images. The final model consisted of 3 streams of CNN models and the outputs are merged at the final layer. The team highlights that many common techniques used for conventional image recognition tasks, such as dropout or batch normalization, were not useful for this particular task. The team placed 2nd having RMSE score of 0.056 in the private leaderboard. 
\subsubsection{Epoch.} The team averaged the steering angles within each timestamp. They also emphasize that k-fold cross validation significantly boosted their performance. Additionally, the team also used a combination of different augmentations, including small rotations, horizontal flips and random brightness, which improved the performance considerably. Interestingly, the team also concludes that smoothing, with the factor of 0.4, to the steering angles output from model improved their final score. 
\subsubsection{Komanda.} Unlike two previous teams, Komanda utilized a 3D CNN-LSTM based architecture, where the discrete time axis is interpreted as the first "depth" dimension allowing the model to learn motion detectors and understand the dynamics of driving. In addition to steering angle, the model also predicts the speed and the torque applied to the steering wheel.
\subsubsection{Autumn.} The unique approach of Autumn was the use of optical flow. Optical flow is the pattern of apparent motion of objects, surfaces, and edges in a visual scene caused by the relative motion between an observer and a scene. However, to take a full advantage of transfer learning, the dense optical flow output was converted from cartesian coordinates to polar, then mapped to the HSV coordinate space. The angular component was mapped to the hue, whereas the magnitude component was mapped to the value. This allows to convert it to RGB or BGR color format. Finally, for the i-th sample they compute k previous optical flow images and take a weighted average, which becomes an input to their CNN architecture. A weighted average allows to focus on  long-term motion, while preventing the need to compute optical flow more than once per frame pair.

%% file: 4_solution.tex
\section{Problem Statement \& Proposed Solution}
\subsection{Problem Statement}
With the introduction of self-driving cars by NVIDIA with the DAVE2 model, many other people have started to add on their knowledge and improve it year by year.
Since the future of self-driving car is quickly coming to a reality with the introduction of AI, we are creating a new AI model to improve current models. 
We are creating a model that will improve over the top placed team in the Udacity challenge.
\subsection{Augmentation}
\begin{figure}
\includegraphics[width=\linewidth]{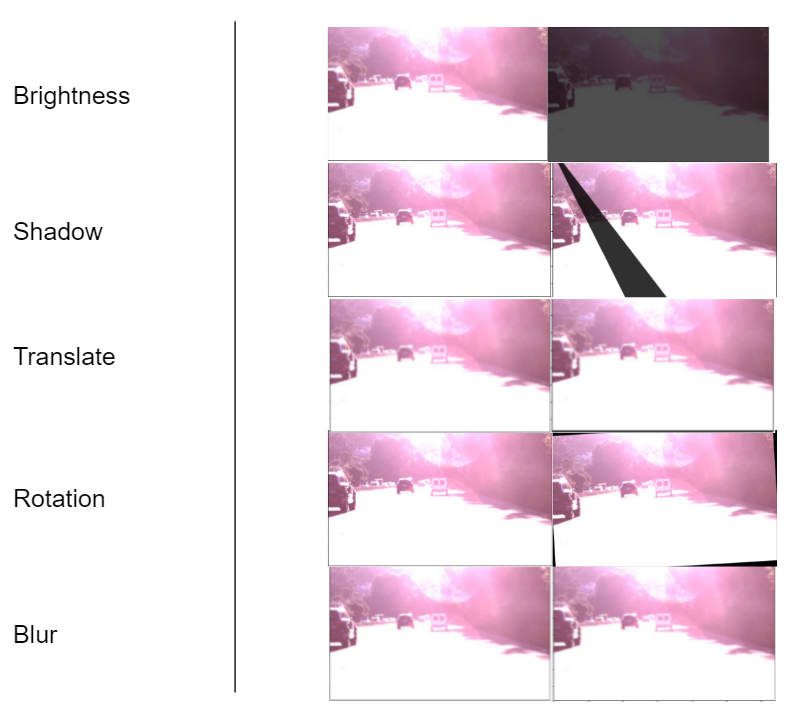}
\caption{Augmentations Used}
\end{figure}
For our dataset we added many Augmentations to help with noise and help our model learn more efficiently. All of our different augmentations can be seen in figure 4.
The first augmentation is brightness. We changed the brightness randomly by converting images to HSV then to RGB. We can clearly see how the brightness was adjusted. The bright pink-like color coming from the reflection of the sun was fierce in the first image, but after the augmentation, the brightness dimmed out and the sun was no longer a factor in the image.
The second augmentation we applied is shadow. We randomly cast shadows to all images by randomly choosing points and shading points on the opposite of it. We can see a shadow cast diagonally in the middle of the screen. This was randomly chosen.
The third augmentation is x and y translations. We do this to add different angles of the road so if the image is left, center or right, we will alter the image to make sure different angles are known. We can see that the image moved up and to the left ever so slightly. This gives images completely new viewpoints, so that no matter from how high or low the image was taken, it will be okay.
The fourth augmentation is rotation. We rotated the images to reduce over-fitting. We apply slight rotations along the center that do not affect the steering angle. We can see that an image is slightly rotated to the left slightly.
The fifth and final augmentation is blur. We blur the images slightly to make sure that even in foggy or rainy conditions, when the windshields are hard to see, the model can still learn effectively for more real life situations.

We observed that weak augmentations worked much better than strong augmentations. Particularly, when we tried to apply stronger augmentations, such as shifting images 30-60 pixels horizontally and 25 pixels vertically, the models tend to simply predict a value close to 0 to all frames. However, when we applied weaker augmentations, such as shifting images 5-10 pixels horizontally and vertically, the models converged much more easily.  
\subsection{Methods}
\begin{figure}
    \centering
    \includegraphics[width=5cm]{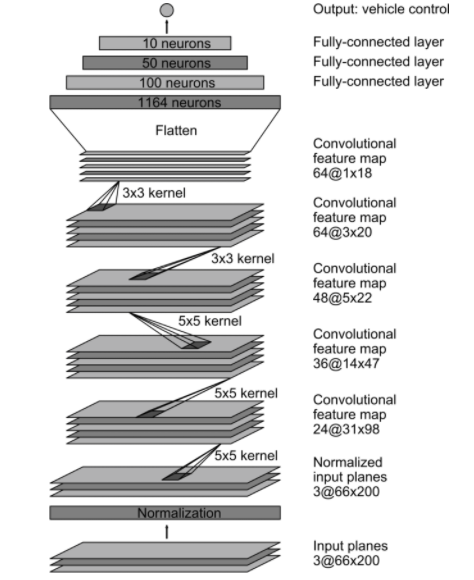}
    \caption{DAVE2 Architecture}
    \label{fig:DAVE2}
\end{figure}
\subsubsection{NVIDIA's DAVE2}
NVIDIA used convolutional neural networks (CNNs) to create a self-steering commands for a self-driving car.
The DAVE2 consists of 9 layers: a normalization layer, 5 convolutional layers, and 3 fully connected layers.
The first layer is responsible for image normalization. This allows normalization scheme to be altered and accelerated by GPU.
The convolutional layers are used for feature extraction.
In the first three convolutional layers, they used strided convolutions with a 2x2 stride and a 5x5 kernel.
In the last two convolutional layers, they use a non-strided convolution with a 3x3 kernel size.
The final three fully connected layers are used as a controller for steering.
For their data selection, they collected data with road type, weather condition, driver's activity and lane following.
They also augmented their images by adding shifts and rotations randomly.
\begin{figure}[h!]
    \centering
    \includegraphics[width=4cm]{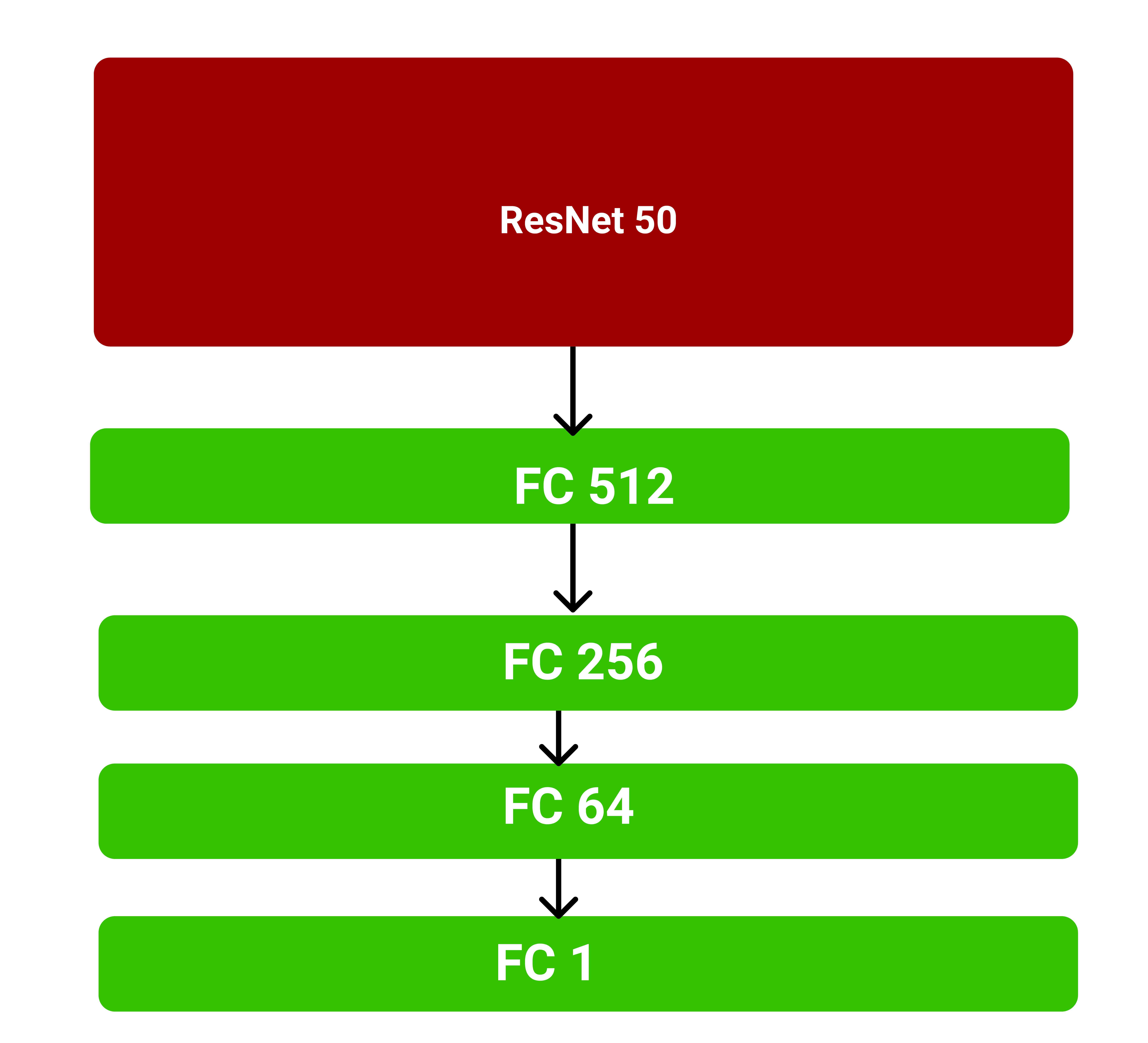}
    \caption{Transfer Learning Architecture}
    \label{fig:transfer}
\end{figure}

\subsubsection{Transfer Learning Model}
This model used a special idea of transfer learning. It is a way of using high quality models that were trained on huge datasets. Features learned in lower layers of model are transferable to another dataset. This helps in areas like edges in the new dataset.
The pre-trained model that they chose was ResNet50 pre-trained on ImageNet. This is because of the pre-trained models available, this one had a good performance on the dataset provided. 
The input into the new model was an image of size 3x224x224. 
After ResNet50, the fully connected layers used ReLUs as activation.
This model proved to be effective in a competitive model and outperformed majority of the other models.
\begin{figure}
    \centering
    \includegraphics[width=4cm]{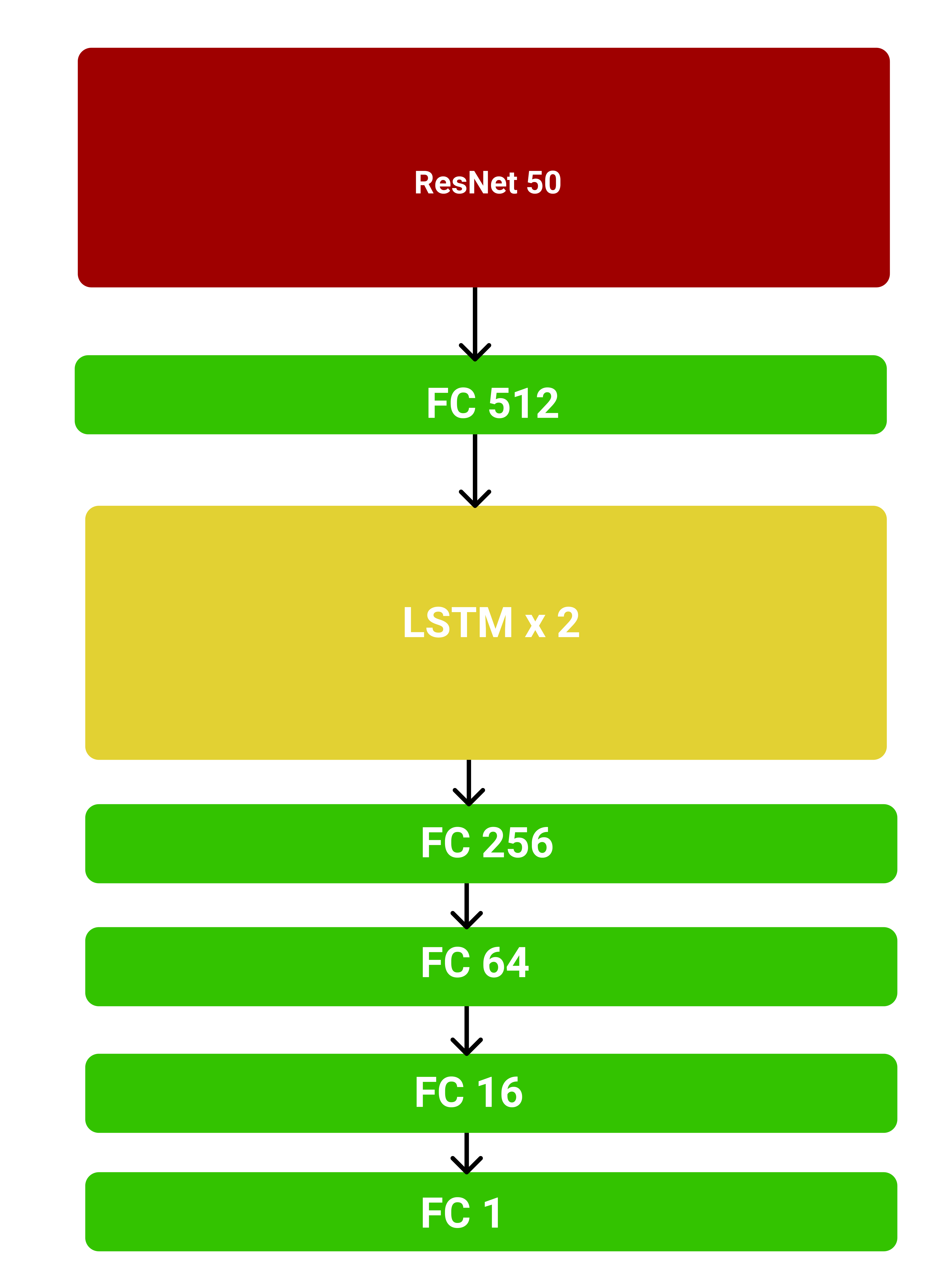}
    \caption{Convolutional Model with Residual Connections and Recurrent LSTM Layers Architecture}
    \label{fig:LSTM}
\end{figure}
\subsubsection{Convolutional Model and Recurrent LSTM Layers}
The input into the backbone, ResNet 50, was 65x3x224x224 in order to take advantage of transfer learning. As we extract the features, we reshape them into 13x5x512, where 5 indicates the sequence length, to make input into LSTM layers smaller. This part consists of 2 layers of LSTM, whereas the output tensor is in the shape of 13x5x256.
Then output of recurrent layer was fed into fully connected stack that outputs an angle prediction. 
All layers mentioned above used rectified linear units (ReLUs) as activation except LSTM.
LSTM Layers used hyperbolic tangent function as activation.
After completing this model, their afterthoughts were that it could definitely become improved upon much further as there was potential for more growth.
\subsection{Discussion}
NVIDIA's DAVE2 architecture was first introduced and tested in real-life scenarios in 2016. Furthermore, it is still relevant and included in many works. In fact, we may observe in few teams' solutions for Udacity Self-driving Car Challenge 2. DAVE2 is a very straightforward and simple architecture; therefore, there have also been efforts to utilize the benefits of transfer learning in steering angle prediction task. Thus, our second baseline architecture adopts a pretrained ResNet50 architecture as a backbone. Although these approaches are able to display quite good results on the dataset, considering that they only take a single frame without a knowledge of the past, we argue and hypothesize that they become biased to an average speed of a car within the dataset. Therefore, they fail to make an accurate prediction in scenarios where a car maneuvers with an accelerating velocity.

In contrast, however, the CNN-LSTM based architecture utilizes a sequence of frames' representations to make the predictions. Due to it, it is able to make predictions utilizing the past information, which represents a car's displacement. Although it takes advantage of a car's past positions, we argue that it lacks explainability and might not be able to extract motion information, which represents how rapidly a car maneuvers along the road. To be more precise, we hypothesize that it makes predictions mainly out of displacement information, which is a change of a car's relative position, since it is not encouraged to learn the change in velocity. We believe that a car's velocity plays a crucial role in predicting future steering angles, especially in scenarios when it has to maneuver under a high velocity. Thus, both displacement and motion information should be taken into account.

In order to tackle this problem, we propose our own approach which encourages to learn both motion and displacement information out of feature representations. Initially, we would like to recall Autumn's, one of the teams mentioned earlier, solution for Udacity Self-driving Car Challenge 2. As we have already emphasized the team utilizes optical flow as input to their network. However, they simply take an average of k-th past optical flow images. We argue that this trivial approach is not always accurate since in real life scenarios certain events should matter more. For example, when a car makes a sudden turn, more recent frames should convey more important information about the trajectory. Moreover, taking the advantage of optical flow images only is not reliable. For example, when a car is on uneven road the camera might change its position horizontally or vertically deceiving the car into believing that it changes its trajectory. Therefore, we utilize both original RGB and optical flow images allowing our network to judge and distribute its attention based on displacement information.  
\subsection{Proposed Solution}
\begin{figure}[h!]
    \centering
    \includegraphics[width=8cm]{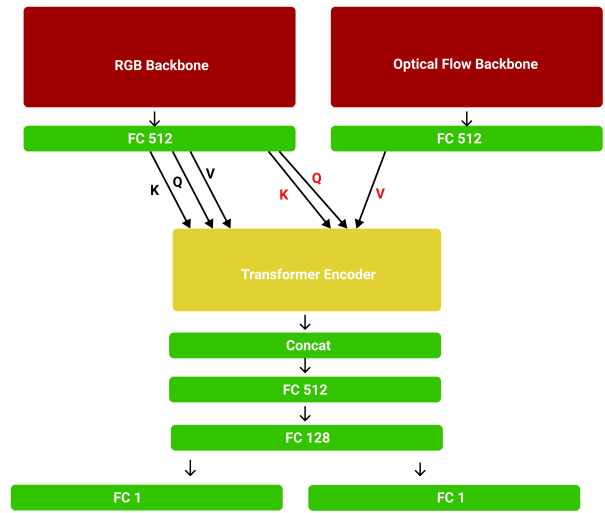}
    \caption{Our proposed architecture.}
    \label{fig:transformer}
\end{figure}
The Figure \ref{fig:transformer} briefly displays our proposed solution. As seen, we utilize two different CNN backbones for RGB and optical flow images. The encoder layer consists of 4 heads and the hidden size is empirically set to 512. Overall, the transformer has 2 layers of encoders. In order to benefit from transfer learning, we adopt ResNet18 for both of the backbones. Thus, we reshape original sequences of $batch\_size \times seq\_len \times 3 \times 224 \times 224$ into $batch\_size * seq\_len \times 3 \times 224 \times 224$. Finally, we reduce the dimensions to 512. Note that the transformer encoder architecture was modified in order to work with the representations coming from two backbones. We expect the representations coming from RGB backbone to convey position information, whereas the ones from optical flow backbone convey motion information. Furthermore, we rely on multi-head attention in stead of RNN layers allowing the model to focus on specific frames' representations to judge a car's trajectory. Likewise, we also use frames' representations as $Key$ and $Query$ parameters for optical flow branch's multi-head attention module. Due to it, we encourage the model to distribute its attention within a sequence of motion representations based on a car's position information, which should be invariant to unanticipated changes in camera's position. We hypothesize that utilizing information in this fashion allows the model's judgement of optical flow representations be robust to unexpected or sudden changes in a camera's position. Additionally, it allows the model to selectively focus on certain frames within the sequence. Next, we concatenate both position and motion embeddings, coming from RGB and optical flow branches respectively, and fuse them using a fully connected layer. Finally, we reduce the dimensions of the fused embedding down to 128 with another fully connected layer. This embedding should contain both displacement and motion information
to make accurate predictions. Thus, we encourage it by predicting both the velocity and steering angle of a car. 

%% file: 4_settings.tex
\section{Training Process}
\subsection{Loss Function and Optimization}
The objective of the Udacity's Self-driving Car Challenge 2 was to optimize fpr RMSE, hence we used it as the training loss function for all of our experiments. The MSE punishes large deviations harshly, whereas RMSE is more suitable for our task. The RMSE loss function can be written as follows:
\begin{align}
    L_{angle} &= \sqrt{\frac{1}{n}\sum(y_{i} - \hat{y}_{i}) ^ 2}
\end{align}
We also predict speed for our proposed architecture only. For this task, we use $L1$ Smooth Loss since it is not our main task. The loss function is written as follows:
\begin{align}
    L_{speed} &= 
\begin{cases}
    0.5 \times \frac{(y_{i} - \hat{y}_{i})^2}{\beta},& \text{if } |y_{i} - \hat{y}_{i}| < \beta\\
    |y_{i} - \hat{y}_{i}| - 0.5 \times \beta,    & \text{otherwise}
\end{cases}
\end{align}
where $\beta$ is a hyperparameter commonly set to $1.0$.\\
To optimize this loss, we use Adam optimizer. The learning rate is set to $0.0001$, $\beta_{1}$ and $\beta_2$ are set to $0.9$ and $0.999$, respectively. Also, the learning rate is decayed by the factor of 10 at 30, 90 and 150 epochs and all the models are trained for 160 epochs. In order to take advantage of transfer learning, we resized the images into 224 by 224; however, DAVE2 utilizes the images of the size of 120 by 320 as in the original paper \cite{DAVE2}. Finally, all the experiments are conducted using Pytorch.  

%% file: 5_Planning.tex
\section{Evaluation}
\begin{figure}[h!]
    \centering
    \includegraphics[width=8cm]{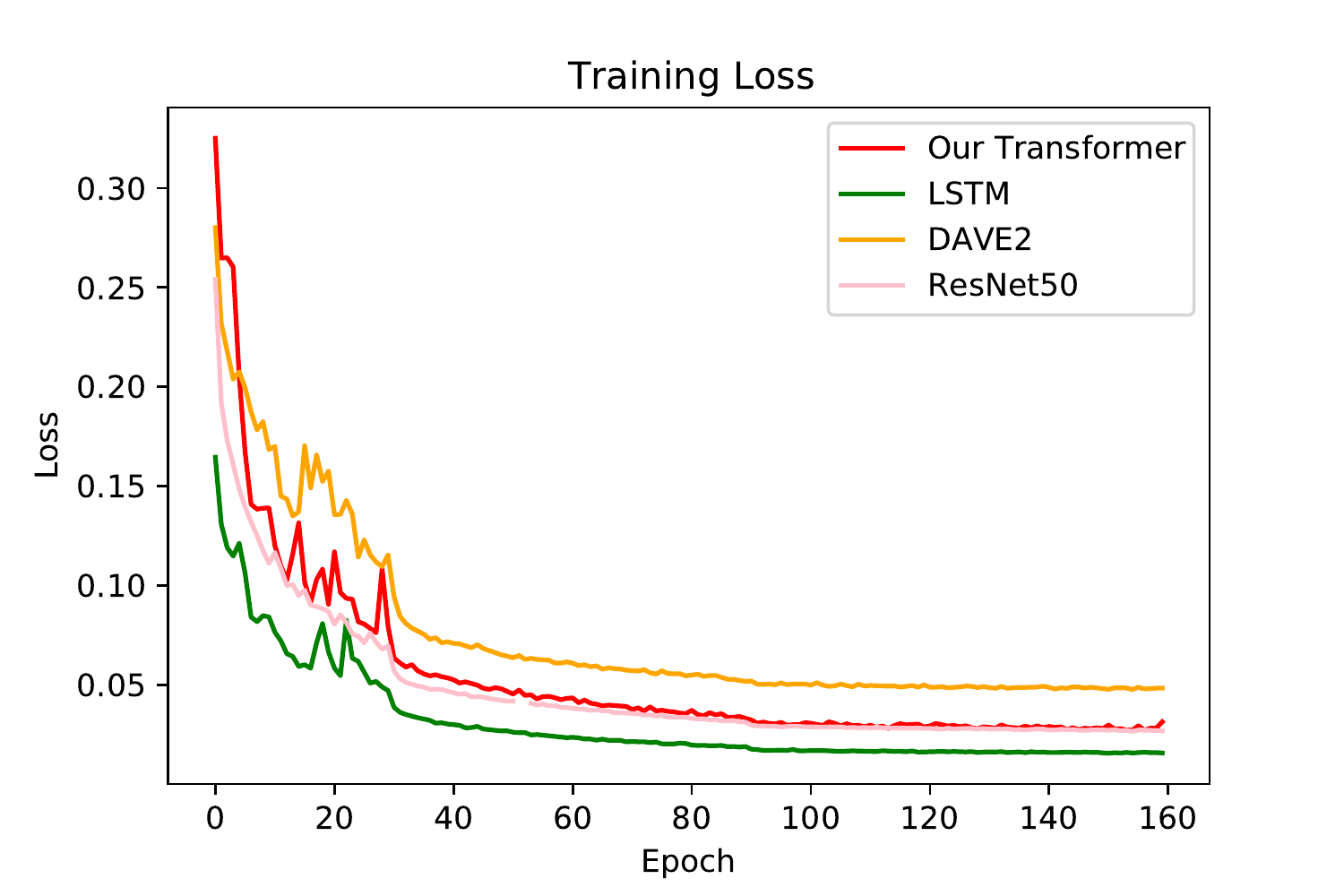}
    \caption{Training curves for all models. The loss function is RMSE.}
    \label{fig:training_main}
\end{figure}
\subsection{Results}
\begin{figure}
\begin{minipage}[c]{0.45\linewidth}
\includegraphics[width=\linewidth]{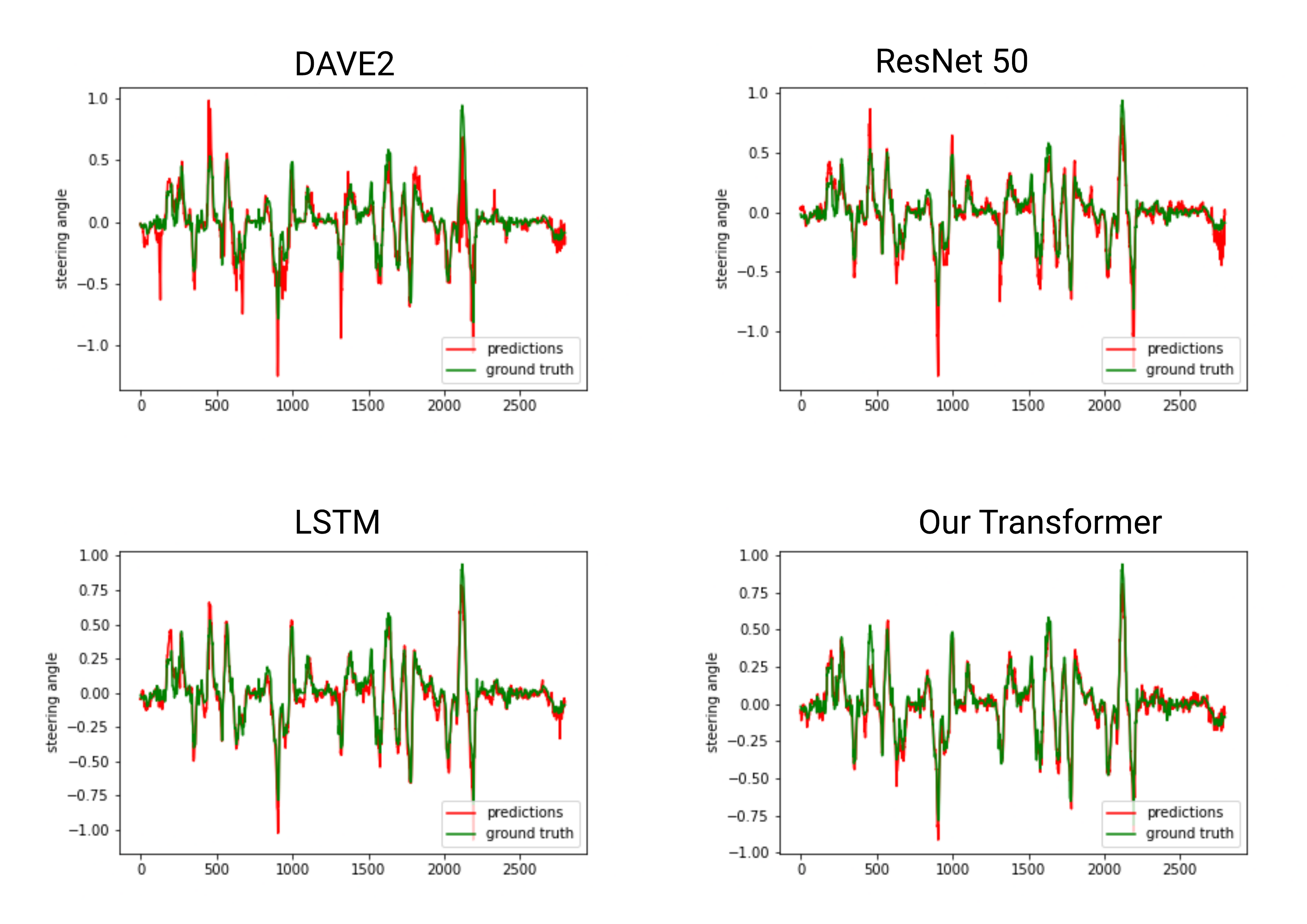}
\caption{Predicted steering angles on public data.}
\label{fig:public}
\end{minipage}
\hfill
\begin{minipage}[c]{0.45\linewidth}
\includegraphics[width=\linewidth]{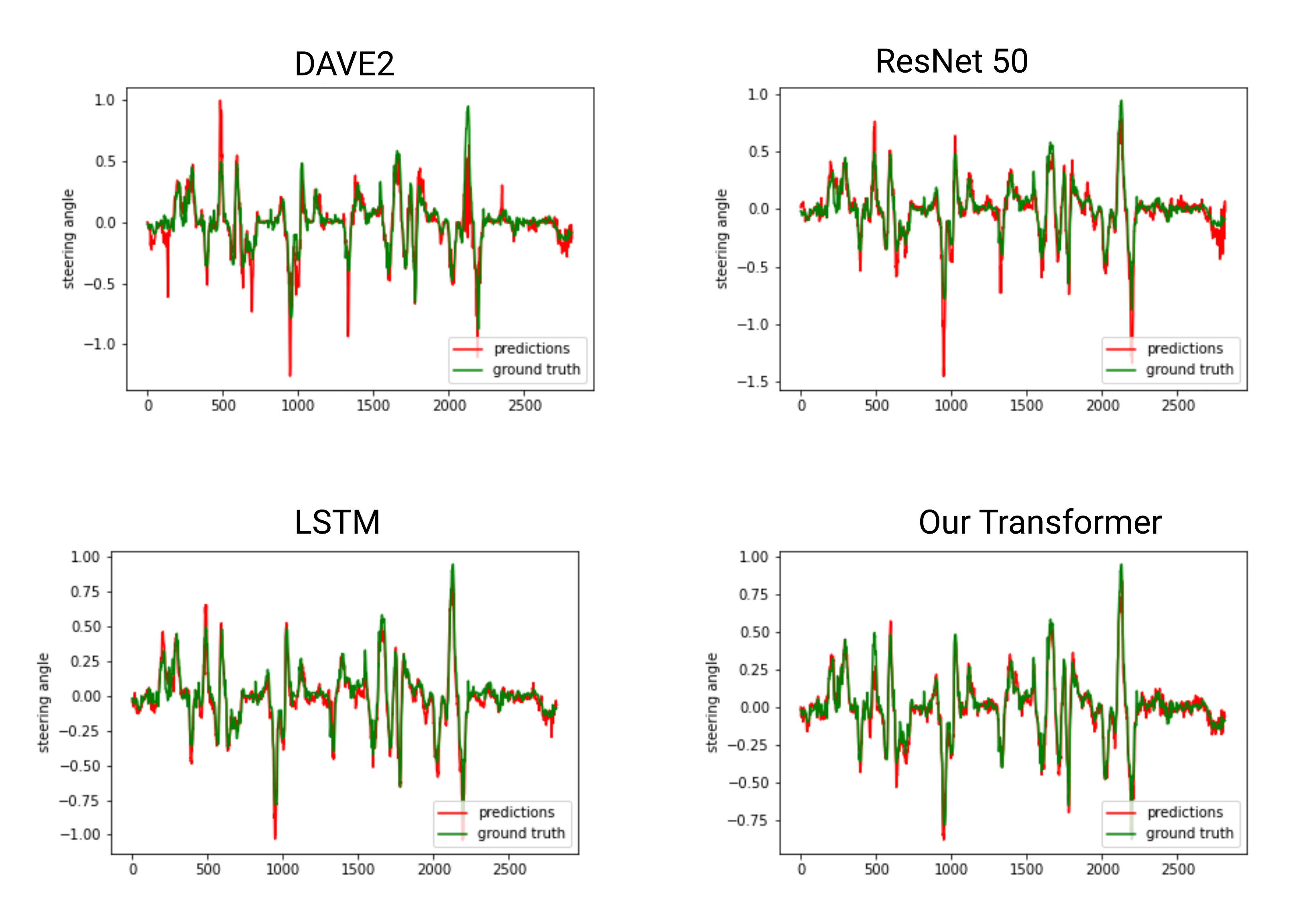}
\caption{Predicted steering angles on private data.}
\label{fig:private}
\end{minipage}%
\end{figure}
The Fig.\ref{fig:training_main} demonstrates the training curves for all the experiments. As we observe LSTM-based model shows the lowest loss figure, whereas DAVE2 has the greatest loss figure. Our proposed models displays a high loss figure at the beginning of the training procedure but converges at a much lower value than DAVE2.
The Fig.\ref{fig:public} shows the models' predictions on the public test data. The read line represents the predictions, whereas target values are represented by the green line. We clearly see that our proposed architecture better handles some extreme values comparing to trivial solutions, such as DAVE2. We argue that the models become biased to the average speed of the car; however, it is challenging to infer the speed from a single RGB frame. The sequence-based modes, on the other hand, are clearly better at these scenarios. We observe that both LSTM-based and our proposed architecture achieved much lower loss at predicting these extreme values. We hypothesize that these models do relatively well by taking advantage of the past information. Finally, we observe the exact same pattern in the Fig.\ref{fig:private}, which shows predictions of the private test data. We also provide a demo video, which is uploaded on YouTube\footnote{https://youtu.be/NeJRwUgqKdQ}, that shows how our proposed model performs on the whole test data.

\begin{table}[ht]
\caption{RMSE for the models on the Udacity dataset. * indicates that the smoothing with the factor of 0.35 is applied.} 
\centering
\begin{tabular}{c c c c} 
\hline\hline 
N & Method & Private & Public \\ [0.5ex] 

\hline
1 & DAVE2 & 0.1327 & 0.1347 \\
2 & ResNet50 & 0.0978 & 0.0981 \\
3 & CNN-LSTM & 0.0718 & 0.0741 \\
4 & Our Transformer & 0.0614 & 0.0631 \\
5 & Our Transformer* & \textbf{0.0577} & \textbf{0.0588} \\ [1ex]
\hline 
\end{tabular}
\label{table:eval} 
\end{table}
The Table  \ref{table:eval} summarizes the performance of the models on both public and private test data. We clearly observe that sequence-based models outperform basic CNN-based approaches. Moreover, our proposed solution outperforms its counterparts by a large margin. We also applied smoothing on the predictions as \textbf{Epoch} and also observe the boost in performance. Our best results are $0.0577$ and $0.0588$ on private and public test data, respectively. The final leaderboard of the Udacity's Self-driving Car Challenge 2 is available online on GitHub\footnote{https://github.com/udacity/self-driving-car/tree/master/challenges/challenge-2}. To sum up, our solution corresponds to the 3rd or 4th place of the challenge. Next, transfer learning- and LSTM-based approaches correspond to the 8th and 6th places of the competition, respectively. Meanwhile, the first place solution is proposed by \textbf{Komanda} which achieved the scores of $0.0483$ and $0.0512$ on public and private test data, respectively. 
\subsection{Ablation study}
Next, we perform an ablation study to demonstrate that additional branch for Optical Flow images contributes to the final performance of our model. Thus, we train another Transformer-based model where we remove the Optical Flow branch. Likewise, we remove the additional task of predicting the speed of the car. We leave other training settings and the backbone for RGB images, ResNet18, unchanged. 
\begin{figure}[h!]
    \centering
    \includegraphics[width=8cm]{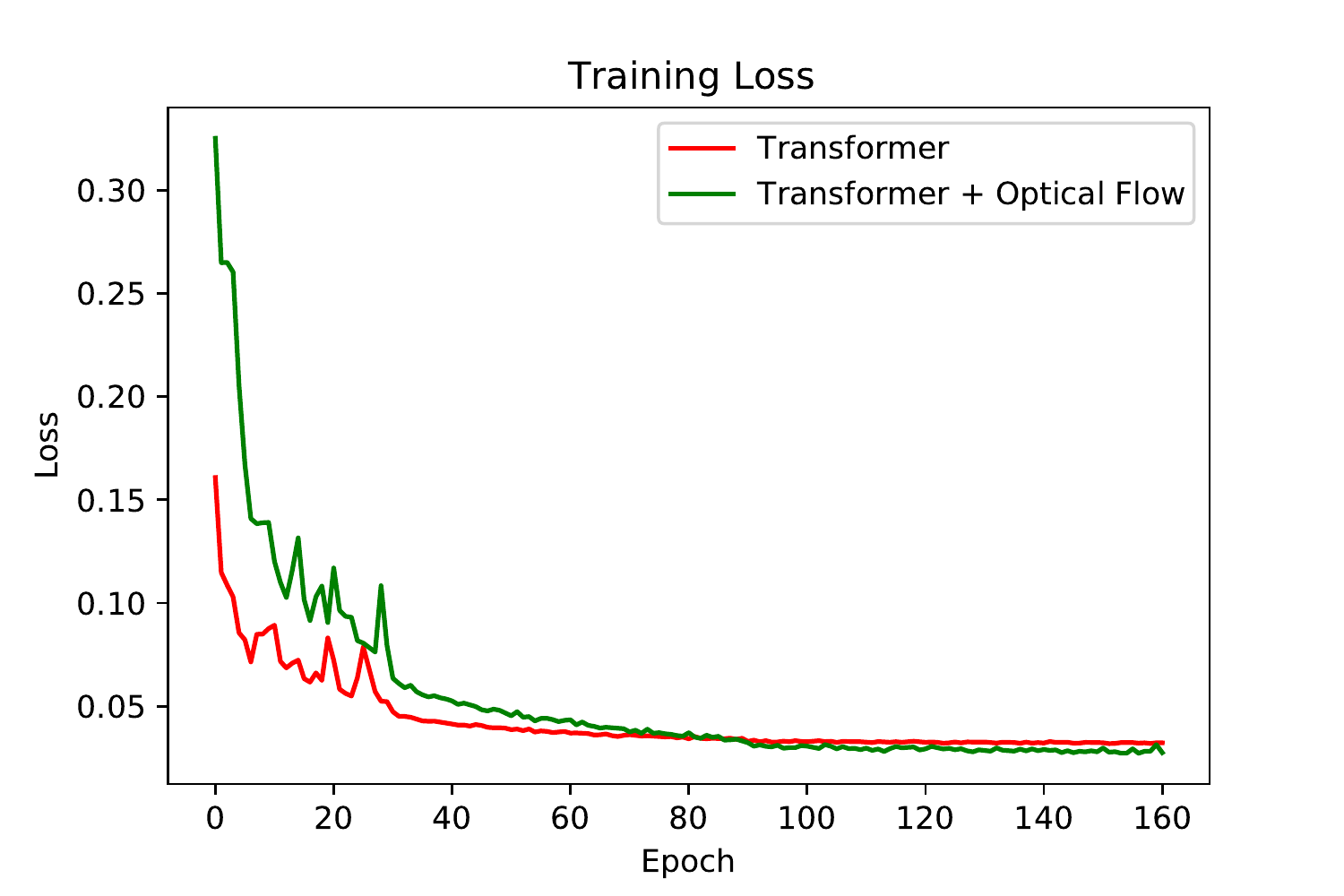}
    \caption{Training curves for both models. The loss function is RMSE.}
    \label{fig:training}
\end{figure}
The Fig.\ref{fig:training} shows training curves for both models. We clearly observe that the training loss for our proposed model was much larger at the beginning of the training procedure; however, both models converged at approximately same loss value.  
\begin{table}[ht]
\caption{RMSE for Transformer models on the Udacity dataset. * indicates that the smoothing with the factor of 0.35 is applied.} 
\centering
\begin{tabular}{c c c c} 
\hline\hline 
N & Method & Private & Public \\ [0.5ex] 

\hline
1 & Simple Transformer & 0.0696 & 0.0706 \\
2 & Our Transformer & 0.0614 & 0.0631 \\
3 & Our Transformer* & \textbf{0.0577} & \textbf{0.0588} \\ [1ex]
\hline 
\end{tabular}
\label{table:eval_transformer} 
\end{table}
The Table \ref{table:eval_transformer} shows that our proposed model outperforms a simple Transformer-based architecture by a decent margin. However, simple Transformer model still outperforms CNN-LSTM based approach. Therefore, we conclude that both branches equally contribute to the performance of our proposed architecture.
\begin{figure}[h!]
    \includegraphics[width=1.\textwidth]{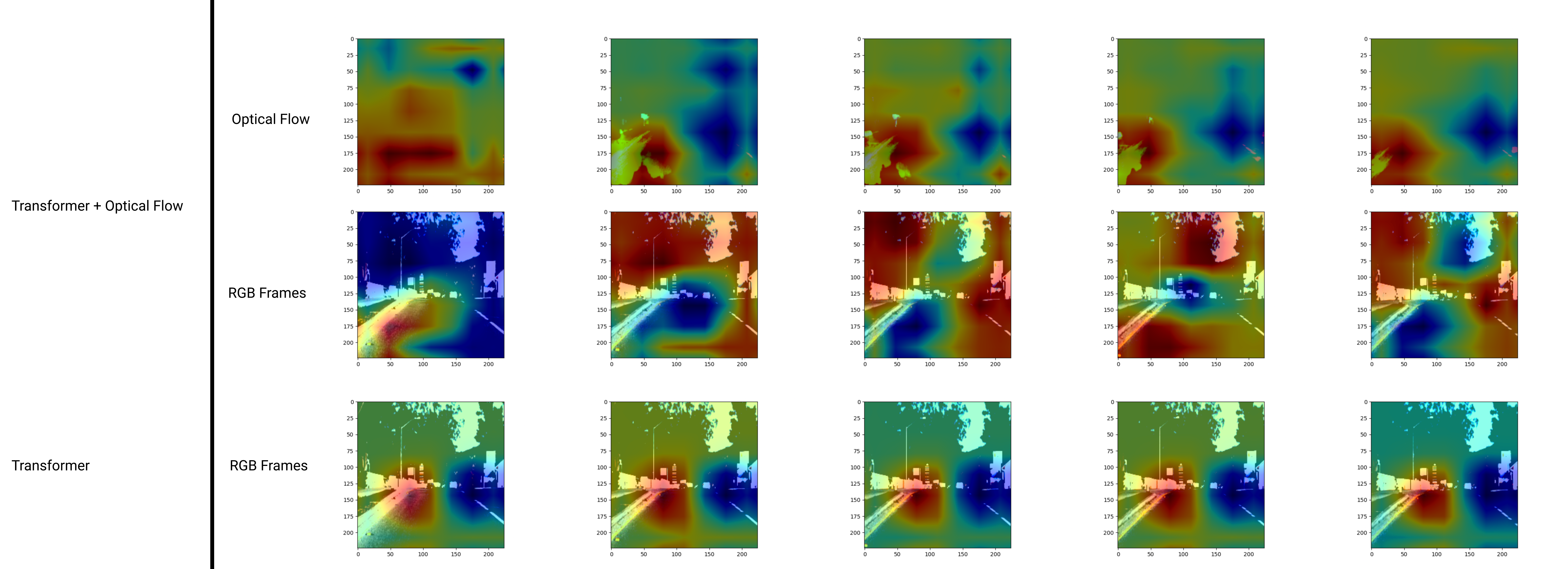}
    \caption{Attention maps for our proposed and simple Transformer model.}
    \label{fig:attentions}
\end{figure}
The Fig.\ref{fig:attentions} shows attentions maps for both transformer and our proposed model. As we observe attention maps for the simple transformer model do not vary significantly over the sequence. In fact, we see that it is mostly focused on the left side of the road. In stark contrast, however, the proposed model seems to vary its attention. One of the possible reasons might be the effect of optical flow images. The first row demonstrates attention maps for optical flow images, where we see the model pay attention to the most significant changes between two consecutive frames (represented within optical flow images).   

%% file: conclusion.tex
\section{Conclusion and Future Work}
In our work, we made an extensive study on Udacity's Self-driving Car Challenge 2. We revisited a list of previous solutions of the teams that participated in the competition, namely Rambo, Epoch, Komanda and Autumn. Furthermore, we introduced other baseline approaches, including DAVE2, ResNet50 and CNN-LSTM based approaches. Moreover, we propose our new architecture to tackle this problem, which was inspired by previous teams' solutions. Our proposed architecture outperforms our baseline architectures by a decent margin achieving $0.0577$ and $0.0588$ on private and public test data, respectively. We also provided an ablation study to demonstrate the efficiency of our solution. Although the results do not outperform the first place solution, we believe that further work can be done to improve our solution. For example, different backbone architectures or loss functions can be used. However, due to time limitations, we leave it for future studies. 

%% file: main.bbl
\begin{thebibliography}{8}
\bibitem{DAVE2}
M. Bojarski, D. Del Testa, D. Dworakowski, B. Firner,
B. Flepp, P. Goyal, L. D. Jackel, M. Monfort, U. Muller,
J. Zhang, et al. End to end learning for self-driving cars.
\emph{arXiv preprint arXiv:1604.07316}, 2016.
\bibitem{RL_car}
A. El Sallab, M. Abdou, E. Perot, and S. Yogamani. Deep
reinforcement learning framework for autonomous driving.
Autonomous Vehicles and Machines, Electronic Imaging,
2017.
\bibitem{scene_parsing}
C. Farabet, C. Couprie, L. Najman, and Y. LeCun. Learning
hierarchical features for scene labeling. \emph{IEEE transactions
on pattern analysis and machine intelligence}, 35(8):1915–
1929, 2013.
\bibitem{GAN}
I. Goodfellow, J. Pouget-Abadie, M. Mirza, B. Xu,
D. Warde-Farley, S. Ozair, A. Courville, and Y. Bengio. Generative adversarial nets. In \emph{Advances in neural information
processing systems}, pages 2672–2680, 2014.
\bibitem{video_class}
A. Karpathy, G. Toderici, S. Shetty, T. Leung, R. Sukthankar,
and L. Fei-Fei. Large-scale video classification with convolutional neural networks. In \emph{Proceedings of the IEEE conference on Computer Vision and Pattern Recognition}, pages
1725–1732, 2014.
\bibitem{Atari}
V. Mnih, K. Kavukcuoglu, D. Silver, A. Graves,
I. Antonoglou, D. Wierstra, and M. Riedmiller. Playing atari with deep reinforcement learning. arXiv preprint
arXiv:1312.5602, 2013.
\bibitem{VAE}
 D. P. Kingma and M. Welling. Auto-encoding variational
bayes. \emph{arXiv preprint arXiv:1312.6114}, 2013.
\bibitem{detection}
C. Szegedy, A. Toshev, and D. Erhan. Deep neural networks
for object detection. In \emph{Advances in Neural Information Processing Systems}, pages 2553–2561, 2013.
%Author, F.: Article title. Journal \textbf{2}(5), 99--110 (2016)

%Author, F., Author, S.: Title of a proceedings paper. In: Editor,
%F., Editor, S. (eds.) CONFERENCE 2016, LNCS, vol. 9999, pp. 1--13.
%Springer, Heidelberg (2016). \doi{10.10007/1234567890}

%Author, F., Author, S., Author, T.: Book title. 2nd edn. Publisher,

%Author, A.-B.: Contribution title. In: 9th International Proceedings
%on Proceedings, pp. 1--2. Publisher, Location (2010)

%LNCS Homepage, \url{http://www.springer.com/lncs}. Last accessed 4
\end{thebibliography}
